\documentclass[11pt]{article}

\usepackage[preprint]{acl}

\usepackage{times}
\usepackage{latexsym}

\usepackage[T1]{fontenc}

\usepackage[utf8]{inputenc}

\usepackage{microtype}

\usepackage{inconsolata}

\usepackage{graphicx}

\newtheorem{definition}{Definition}

\usepackage{amsmath}
\usepackage{xcolor}
\usepackage[ruled,vlined,linesnumbered]{algorithm2e}
\usepackage{multirow}
\usepackage{listings}
\usepackage{tablefootnote}

\title{\textsc{DIP:}\xspace Dynamic In-Context Planner For Diffusion Language Models}

\author{Yang Li\textsuperscript{1} \quad Han Meng\textsuperscript{1} \quad Chenan Wang\textsuperscript{1} \quad Zhenyu Bi\textsuperscript{2} \quad Xuan Wang\textsuperscript{2} \quad Haipeng Chen\textsuperscript{1} \\
College of William \& Mary\textsuperscript{1}, Virginia Tech\textsuperscript{2} \\
  \texttt{\{yli102, hmeng, cwang33, hchen23\}@wm.edu\textsuperscript{1}, \{zhenyub, xuanw\}@vt.edu}\textsuperscript{2}}

\begin{document}
\maketitle
\begin{abstract}
Diffusion language models (DLMs) have shown strong potential for general natural language tasks with in-context examples. Existing In-Context Learning (ICL) approaches largely inherit the practice of autoregressive language models (ARLMs), incorporating all examples into a fixed prompt. However, applying this rigid, static-prompt paradigm to DLMs incurs substantial computational overhead, as the model must evaluate the maximum context length at every step. We address this inefficiency with a key discovery: the block-wise KV-cache mechanism inherent to DLM inference enables the \textit{low-cost dynamic adjustment of the context}. Following this intuition, our core idea is to start generation with a minimal prompt and progressively insert additional examples on the fly only when the generated tokens are of low confidence. Through rigorous empirical evaluations, we observe that average verified token confidence correlates strongly with generation accuracy, making it a reliable and computationally efficient signal of token quality.
Formally, we propose \textbf{D}ynamic \textbf{I}n-Context \textbf{P}lanner (DIP), a context-optimization algorithm based on average verified confidence that dynamically ranks and inserts in-context examples during generation, rather than providing all examples up front. Experimental results on math and coding benchmarks with LLaDA-1.5 and LLaDA-8B-Instruct show that DIP achieves up to $1.59\times$ and $1.36\times$ speedups, respectively, while largely preserving the generation quality of the fixed-prompt baseline. Code: \url{https://github.com/wmd3i/DIP}

\end{abstract}

\section{Introduction}

Diffusion language models have emerged as a new paradigm for language modeling, demonstrating strong potential across general natural language tasks \citep{nie2025large,ye2025dream,austin2021structured,deepmindGeminiDiffusion,arriola2025block, lou2023discrete,sahoo2024simple}. 
DLMs' decoding mechanism differs fundamentally from that of ARLMs: By utilizing bidirectional attention and iterative refinement, DLMs remove the strict left-to-right constraint inherent to AR generation. 
This paradigm shift offers fine-grained control over the generation process, enabling flexible editing and global planning \citep{zhang2025survey,jin2025thinking,li2025diffusion,chen2025dparallel}.

Similar to ARLMs, DLMs also benefit from In-Context Learning (ICL) \cite{nie2025large}, in which providing in-context examples improves performance on many retrieval-augmented generation tasks \citep{gao2023retrieval}. 
Existing ICL approaches for DLMs largely inherit the AR convention \citep{radford2018improving,bi2025optagent}, which keeps the prompt fixed and uses all in-context examples as the prefix throughout generation \citep{radford2018improving,dong2024survey}.
While this practice is reasonable in ARLMs, the higher computational cost of bidirectional attention in DLMs significantly limits inference speed as the context size grows.

To reduce computational cost in DLMs, recent works have explored KV caching \citep{liu2025dllm, song2025sparse}, parallel decoding \citep{jiang2025d, wu2025fast}, and block-wise decoding \citep{wu2025fast}. 
While these approaches successfully mitigate the overhead of processing static prompts, they overlook a critical source of computational waste: the rigid assumption that the model requires a maximum fixed context length throughout the generative process. 

\begin{figure*}[t]
\centering
  \includegraphics[width=\textwidth]{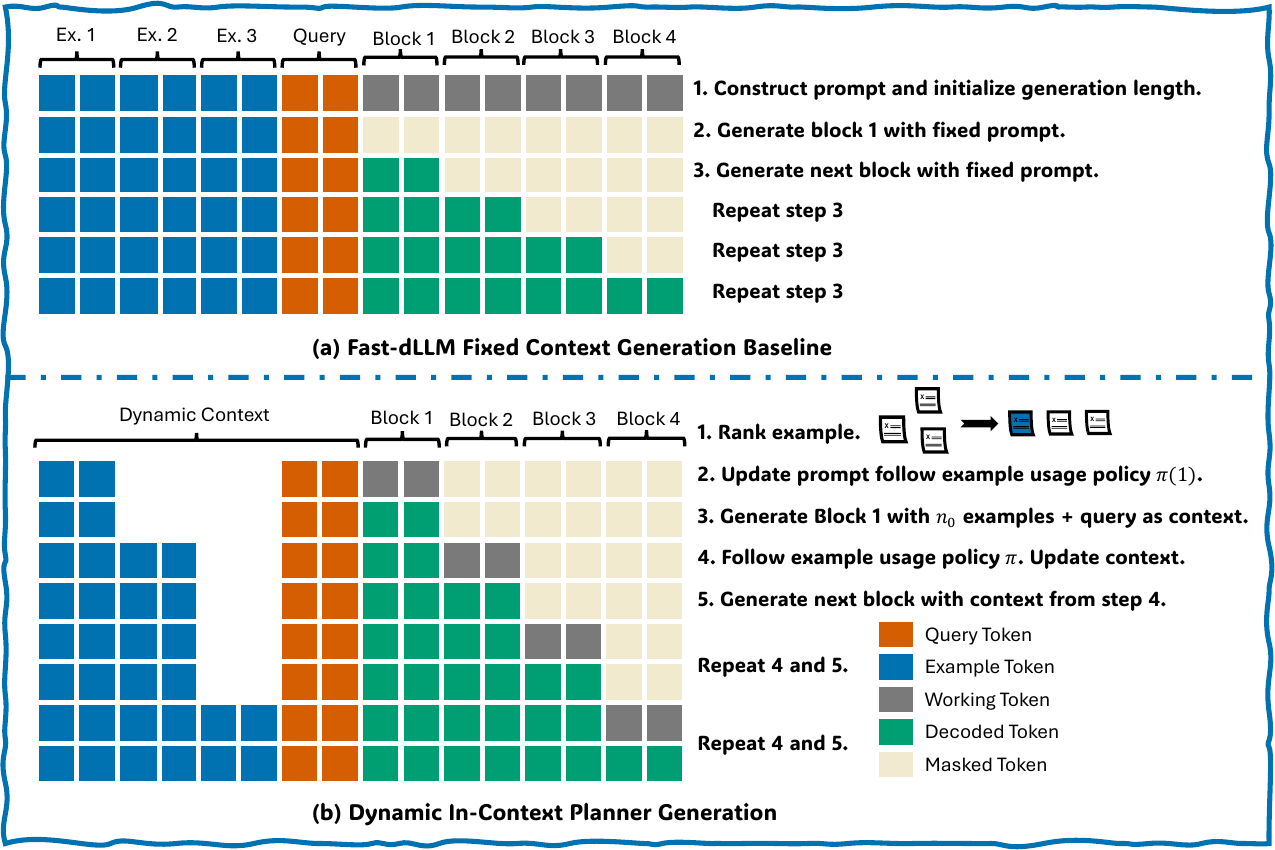}
  \caption{\label{fig:DIP} Inference process of fixed context generation baseline and dynamic in-context planner inference using Fast-dLLM \cite{wu2025fast} inference framework. (a) Illustration of the Fast-dLLM generation process under a fixed context setting, which is commonly used in 3-shot in-context learning scenarios. (b) Illustration of a dynamic context policy $\pi$ based generation process. During decoding, each block may use a different number of in-context examples, depending on the policy. A more detailed example can be found in Appendix \ref{apd:example}.
  }
 
\end{figure*}

This work addresses this issue with a \textbf{key discovery}: unlike ARLMs, which require a fixed context, the block-wise KV-cache mechanism inherent in DLM inference enables the \emph{dynamic adjustment of context} across decoding blocks at low computational cost. Exploiting this mechanism, our core idea is to start generation with a shorter prompt and progressively increase the context length on the fly, thereby significantly reducing the required attention computation without sacrificing quality. To make this concrete, consider a few-shot GSM8K task. In ARLMs, all examples must be rigidly loaded into the prompt before generation begins. 
By contrast, as illustrated in Fig. \ref{fig:DIP} part (b), the block-wise KV cache enhanced diffusion generation paradigm allows for generation with varying numbers of examples across different stages of the process 
(e.g., 1-shot at block 1, 2-shot at block 2 and 3, and 3-shot at block 4). By dynamically adapting the number of in-context examples to the evolving needs of the generation, we can significantly reduce computational cost in earlier decoding stages while leveraging the full set of in-context examples in later stages, without compromising generation quality.


A critical question naturally emerges: on what basis should we dynamically adjust the context? To fully exploit this architectural freedom, we need a reliable, real-time proxy to tell us when the model is struggling.
Motivated by recent work on predictive metrics and test-time methods \cite{fu2026deep, li2026beyond, geng-etal-2024-survey}, we investigated the interplay among predictive metrics, task accuracy, and the number of in-context examples. 
Our analysis revealed clear empirical patterns of the average verified confidence for the generated tokens, making it an ideal signal for our selection policy: (1) correct and incorrect answers exhibit a distinct distributional difference in the average verified confidence of the generated tokens (Fig. \ref{fig:Conf-Corr}); (2) scaling up the number of in-context examples reliably boosts confidence and accuracy (Fig. \ref{fig:N-ACC}).

Building on these insights, we propose \textbf{D}ynamic \textbf{I}n-Context \textbf{P}lanner (DIP), an average verified confidence-based context-optimization algorithm that dynamically selects in-context examples during generation.
DIP consists of a sentence embedding-based example-ranking stage and an average verified confidence-aware example-insertion policy that makes insertion decisions during generation. 
DIP is training-free and can serve as a plug-in method that efficiently selects examples and updates the prompt using prominent DLM inference frameworks like Fast-dLLM \cite{wu2025fast}.

Our main contributions include: 
(1) We demonstrate that DLMs' generation paradigm allows \textit{low-cost dynamic context adjustment}, breaking the rigid static-prompt assumption of traditional ARLMs, opening a new avenue for efficient DLMs inference based on dynamic context optimization.
(2) We investigate DLMs' predictive metrics, 
establishing that average verified confidence could serve as an accurate, real-time signal for when to inject new examples.
(3) We introduce DIP, a practical context optimization framework that dynamically scales in-context examples during generation based on verified confidence.
(4) Experimental results on math and coding benchmarks with LLaDA-1.5 and LLaDA-8B-Instruct show that DIP achieves up to $1.59\times$ and $1.36\times$ speedups, respectively, while largely preserving the generation quality of the fixed-prompt baseline.

\section{Related Work}
\noindent\textbf{In-Context Learning.} ICL enables language models to perform tasks by conditioning on a few input-output demonstrations in the prompt without parameter updates~\citep{brown2020language, dong2024survey}. 
Prior work has shown that the choice of demonstrations significantly affects performance: 
\citet{min2022rethinking} find that format and input distribution matter more than label correctness, while \citet{liu2022what} and \citet{rubin2022learning} demonstrate that retrieving semantically similar examples substantially outperforms random selection. 
However, in ARLMs, all examples must be fixed before generation begins. 
Recent DLMs \citep{nie2025large, ye2025dream, arriola2025block} are free from this causal constraint, opening new possibilities for ICL. For instance, 
\citet{jin2025thinking} explore in-place condition injection.  
Yet, all existing approaches still fix the prompt before generation. 

\noindent\textbf{Efficient Inference in DLMs.}
Masked DLMs use bidirectional attention, which typically scales quadratically with the input and generation lengths. Following the success of KV cache reuse in ARLMs \cite{kwon2023efficient,zheng2024sglang}, recent works, like Fast-dLLM \cite{wu2025fast} and dKV-Cache \cite{ma2025dkvcache}, have achieved significant inference speedup with KV cache reuse \cite{jiang2025d,song2025sparse,liu2025dllm,ma2025dkvcache}. 
Other lines of work also explore parallel decoding \cite{wu2025fast, israel2025accelerating,zhang2025survey} to sample multiple tokens per iteration, thereby improving inference speed. 
Despite these initial successes, these methods focus solely on the computational cost of a static prompt and ignore the opportunity to dynamically adjust the context used for generation.

\noindent\textbf{Token Confidence In Language Models.}
Token confidence or logits have gained increasing attention in language model research. They are not only useful for token generation but also reveal many insights into the generated tokens. In ARLMs, recent work has used reasoning-trace-level confidence to enable effective test-time scaling and trace selection \cite{fu2026deep, kang2025scalable}. Token confidence in DLMs could also reveal insights, such as whether the generation length is sufficient based on the EOS token's confidence \cite{li2026beyond}. Despite these initial successes, the correlation between DLM's trace-level confidence and the number of in-context examples remains poorly understood. 

We aim to fill these research gaps by exploring DLMs' token confidence and dynamically adjusting in-context examples during diffusion generation, exploiting the unique flexibility of DLMs' bidirectional attention.

\section{Preliminary: Masked DLMs}\label{Prelim}
Masked DLMs (absorbing state discrete diffusion models) \citep{sahoo2024simple,zhu2025llada,ye2025dream, austin2021structured} consist of a forward and reverse process, where an absorbing state is a mask token, denoted as [MASK], which represents missing or corrupted information \citep{lou2023discrete, nie2025large}.


In the \textit{forward} process, $q$ with an initial sequence $y_0$ of length $N$ is organized into discrete \textit{time steps}, where each time step corresponds to a single masking stage. At each time step, independent tokens are gradually replaced by [MASK] tokens with probability $t\in[0,1]$, which is defined as:
\begin{equation}
    q(y_t|y_0) = \prod\nolimits^{N-1}_{i=0} q(y_t^i|y_0^i),
\end{equation} where $i \in [0,N]$ and $y^i$ stands for the $i$-th token. Once a token transitions to [MASK], it remains in this state throughout the forward process. As $t$ increases to $1$, the original sequence will eventually become a fully masked sequence. 

Conversely, in the \textit{reverse} process $p_\theta$, at each time step, masked DLMs aim to predict the probability distribution over all masked tokens simultaneously given $y_t$, which can be formulated as:
\begin{equation}
    p_\theta (y_s|y_t) = \prod\nolimits_{i=0}^{N-1}p_\theta(y_s^i|y_t^i).
\end{equation}
Based on these predictions, the additional decoding method $D$ decides which tokens to "unmask" at each step. As $t$ decreases to $0$, the masked sequence will eventually recover the original sequence.



\section{Accelerating DLMs via Dynamic Context Adaptation}

\noindent\textbf{ARLMs via Causal Attention.}
ARLMs \cite{hu2024improving} employ causal attention~\citep{vaswani2017attention} for sequential generation, restricting the contextualized representation $h_i$ to left-to-right dependencies ~\citep{radford2018improving, brown2020language}. This causal masking can be formally expressed by bounding key and value indices to $\le i$:
\begin{equation}\label{eq:causal_attn}
    h_i = \text{Attention}(x_i W_Q, \, x_{\le i} W_K, \, x_{\le i} W_V ) \; .
\end{equation}
Because the structural context is rigidly bound to the prefix, the key and value are cached sequentially to avoid redundant computation. 

As a result, ARLMs are fundamentally inefficient in dynamic context scenarios during inference. Any mid-generation modification to the context instantly forces a recomputation of the invalidated sequence. Rebuilding this cache from scratch imposes a computational cost of approximately:
\begin{equation}\label{eq:full_length}
    O((\text{L}_{new} + \text{L}_{gen})^2) \; ,
\end{equation}
where $\text{L}_{new}$ is the length of the new context and $\text{L}_{gen}$ is the length of generated tokens so far. 

\noindent\textbf{DLMs via Full Attention.}
In contrast, masked DLMs \cite{zhu2025llada, ye2025dream} employ full attention~\citep{devlin2019bert}, 
where the $i$-th token attends to the entire sequence of length $L$ to compute its representation $h_i$:
\begin{equation}\label{eq:full_attn}
    h_i = \text{Attention}(x_i W_Q, \, x_{1:L} W_K, \, x_{1:L} W_V ) \; .
\end{equation}
During generation, DLMs iteratively refine a fully masked sequence by predicting and unmasking tokens across multiple denoising steps. Because this full attention mechanism inherently recomputes the entire sequence at each step, updating the context mid-generation does not trigger the cache-rebuild penalty observed in ARLMs. Yet, this full attention mechanism in DLMs renders long contexts computationally expensive.


\noindent\textit{\textbf{Key Discovery:} Low-cost dynamic context adaptation in DLMs allows us to actively adjust the context mid-generation, achieving substantial computational savings.}
In practice, recent KV cache inference frameworks such as Fast-dLLM~\citep{wu2025fast} adopt block-wise generation\footnote{While we focus on block-wise DLM inference as it is the predominant inference framework, this discovery also holds for non-block-based DLM inference.}, which divides the output into $N$ equal-length blocks and performs semi-AR decoding at the block level. Crucially, at each block boundary, a full forward pass is already required to recompute the attention states for the next block. This provides a natural, low-overhead opportunity to update the prompt (e.g., inserting an in-context example) without incurring penalties beyond basic tokenization and matrix concatenation. 

Given a full static prompt containing all in-context examples with context length of $\text{L}_{F}$, and a reduced dynamic prompt with length $\text{L}_{new}$ (where $\text{L}_{new} \le \text{L}_{F}$), one can achieve a significant reduction in the per-denoising iteration cost, which can be roughly estimated as:
\begin{equation}\label{eq:reduction}
    O((\text{L}_{F} + \text{L}_{gen})^2-(\text{L}_{new} + \text{L}_{gen})^2) \; ,
\end{equation}
where $\text{L}_{gen}$ is the length of the generated tokens. Because DLM decoding requires evaluating bidirectional attention across multiple iterative denoising calls per block, maintaining a minimal $\text{L}_{new}$ when the model is confident yields substantial, compounding reductions in overall computation.



\section{Problem Statement}
Following the above key finding, we define a new dynamic in-context planning task in DLMs: instead of using all examples as context throughout the DLM's generation process, we aim to determine when and which examples to insert during generation to accelerate inference while maintaining inference quality in an efficient inference setting.

\begin{definition} \textbf{(ICL Setting)}
    Given a prompt $p$ with a pool of in-context examples $E_{pool}=\{(q_i,a_i)\}_{i=1}^{K}$, where $q_i$ and $a_i$ denote the query and answer of the $i$-th example, respectively.
\end{definition}
\begin{definition} \textbf{(Efficient Inference Setting)}
    Given a KV cache and block diffusion-based inference process divided into $N$ non-KV cache steps (or blocks), indexed by $b\in\{1,\cdots,N\}$.
\end{definition}

\noindent\textbf{Goal:} 
The objective is to find a training-free policy $\pi$ that selects a subset of examples  ${E}_b \subseteq \{{E}_{pool}, \emptyset\}$ for each block $b$, which minimizes the cumulative computational cost while preserving the baseline generation quality.

\section{Dynamic In-Context Planner}
In this section, we formally introduce Dynamic In-Context Planner (DIP). As shown in Figure \ref{fig:DIP}, DIP is a light weight framework that decouples selection complexity into: (1) an example ranking stage, which ranks candidate examples in $E_{pool}$ based on their sentence embedding, and (2) an example insertion policy, which is a monotonic-increasing policy that progressively inserts these examples into the context at different generation blocks $b$ based on generation confidence. Appendix \ref{apd:algo} outlines the formal algorithm.

\begin{figure}[t]
\centering
  \includegraphics[width=\columnwidth]{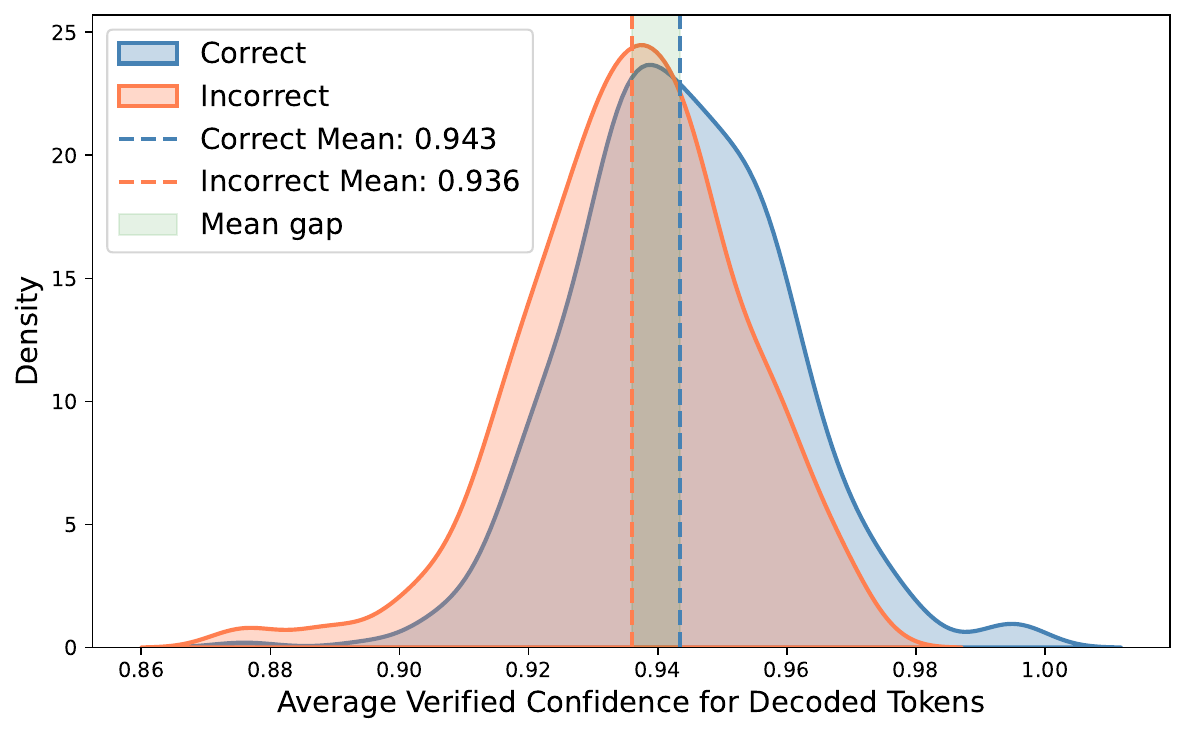}
  \caption{\label{fig:Conf-Corr}Average verified confidence distribution for correct vs. incorrect traces using LLaDA-8B-Instruct on GSM8K. There are 400 data points for each of the correct and incorrect categories. 
  }

\end{figure}

\subsection{Example Ranking}
\begin{figure}[t]
\centering
  \includegraphics[width=\columnwidth]{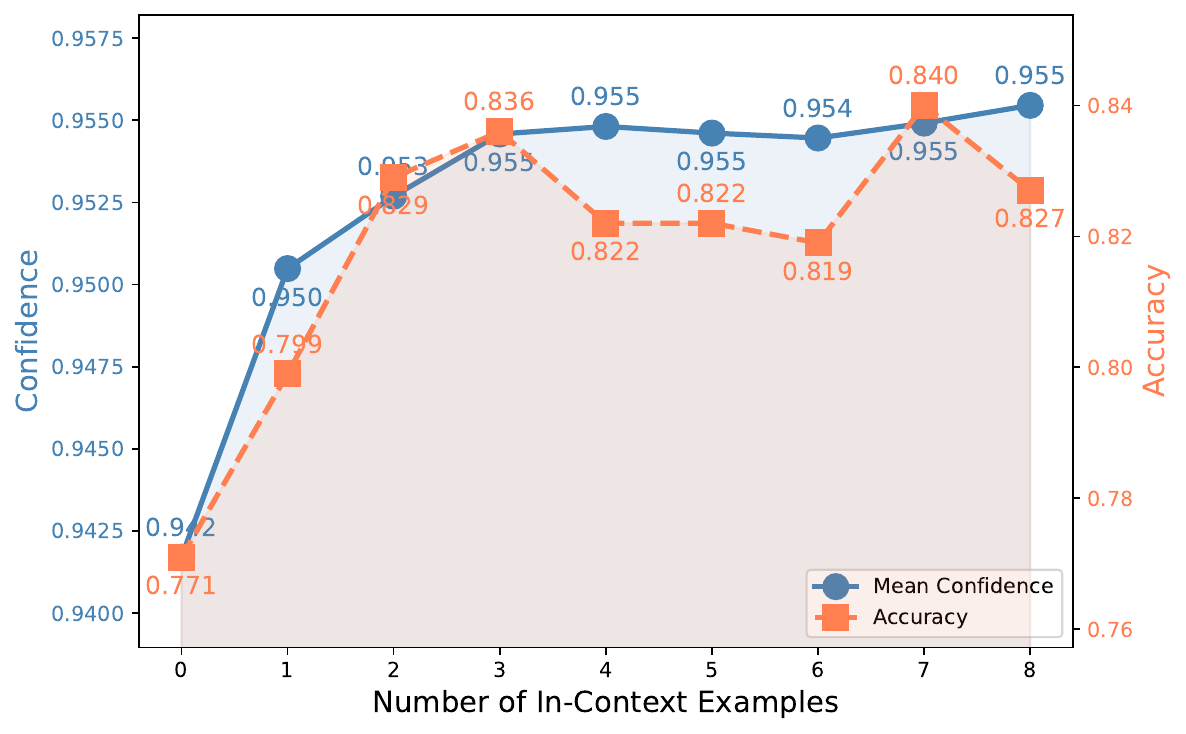}
  \caption{\label{fig:N-ACC} Impact of number of in-context examples vs. confidence and accuracy. Data are collected from the same 1000 sampled questions in the GSM8K training set with 0 to 8 in-context examples.}
\end{figure}

The effectiveness of in-context learning relies heavily on the quality of the examples provided \citep{carbonell1998use,pickett2024better,nafee2025dynamic}. As Fig. \ref{fig:N-ACC} highlights, injecting additional demonstrations can sometimes degrade rather than improve performance. To balance our dual goals of high-speed inference and sustained accuracy, DIP employs a training-free ranking stage to dynamically filter and select only the most beneficial examples during generation.

We tested two common sentence embedding-based \cite{reimers-gurevych-2019-sentence} approaches to rank examples: (1) cosine similarity \cite{zebaze-etal-2025-context, reimers-gurevych-2019-sentence,zhang2025qwen3}, and (2) maximal marginal relevance (MMR) \cite{carbonell1998use,pickett2024better}, which could adjust the relevance and diversity of examples based on the choice of $\lambda$. 
\begin{equation}
\begin{aligned}
    \text{MMR}(q_i, \lambda) &= \; \lambda \cdot \text{Sim}(q_p, q_i) \\
    &\quad -(1-\lambda) \max_{j \in S} \text{Sim}(q_i, q_j)\;,
\end{aligned}
\end{equation} where Sim$(\cdot,\cdot)$ denotes the cosine similarity, $q_p$ is the representation of the actual query the DLM aims to solve, 
$\mathcal{S}$ is the index set of already selected examples,
and $\lambda\in[0,1]$ controls the trade-off between relevance and redundancy. As $\lambda \rightarrow 1$, the ranking favors similarity; as $\lambda \rightarrow 0$, the ranking favors diversity. It is important to note that when $\lambda =1$, MMR is effectively the same as cosine similarity.
\begin{table*}[ht]
\centering
\begin{tabular}{|l|l|l|l|l|l|}

\hline
&Dataset&\multicolumn{2}{c|}{GSM8K (8 shot)}  &\multicolumn{2}{|c|}{Math 500 (8 shot)}\\
\hline
Model & Method     & Throughput & Acc & Throughput & Acc \\ \hline
&Fast-dLLM Fixed Prompt            &        66.58(\textcolor{blue}{1.0$\times$})      & 77.3        &  63.69(\textcolor{blue}{1.0$\times$}) & 38.2  \\
LLaDA-8B&DIP($n_0$=1, RT=0.00)    &     87.80(\textcolor{blue}{1.32$\times$})  &  76.2  &76.34(\textcolor{blue}{1.20$\times$})&32.4
\\
\ \ Instruct&DIP($n_0$=2, RT=0.00)&     \textbf{84.08}(\textcolor{blue}{1.26$\times$})  &  \textbf{78.6}  &\textbf{72.63}(\textcolor{blue}{1.14$\times$}) &\textbf{36.2}\\
&DIP($n_0$=1, RT=0.05)           &     90.24(\textcolor{blue}{1.36$\times$})  &   76.4  &78.13(\textcolor{blue}{1.23$\times$}) &33.2\\
&DIP($n_0$=2, RT=0.05)&   86.49(\textcolor{blue}{1.30$\times$})  
&    77.4     &74.19(\textcolor{blue}{1.16$\times$}) &35.6\\

\hline
&Fast-dLLM Fixed Prompt             &  58.60(\textcolor{blue}{1.0$\times$})      &  80.1  &  49.04(\textcolor{blue}{1.0$\times$}) & 37.2 \\
\multirow{2}{*}{LLaDA 1.5}&DIP($n_0$=1, RT=0.00)   &     87.07(\textcolor{blue}{1.49$\times$})  &   78.9     & \textbf{76.28}(\textcolor{blue}{1.56$\times$})& \textbf{37.2}\\
&DIP($n_0$=2, RT=0.00)&     \textbf{82.92}(\textcolor{blue}{1.42$\times$})  &  \textbf{80.3}   & 73.64(\textcolor{blue}{1.50$\times$})& 36.8\\
&DIP($n_0$=1, RT=0.05)           &     87.73(\textcolor{blue}{1.50$\times$})  &  79.2   & 78.00(\textcolor{blue}{1.59$\times$})& 36.4\\
&DIP($n_0$=2, RT=0.05)&     83.66(\textcolor{blue}{1.43$\times$})  &   80.2    & 74.71(\textcolor{blue}{1.52$\times$})& 36.6\\ \hline
\end{tabular}
\caption{DIP's results on 8-shot GSM8K and 8-shot MATH500 compared against fixed prompt baseline using LLaDA-8B-Instruct \citep{nie2025large} and LLaDA-1.5 \cite{zhu2025llada}. \textcolor{blue}{Blue: relative speedup.} We also \textbf{bold} the DIP configuration that achieves the highest speedup while largely preserving or yielding better generation quality. }
\label{tab:main-8-shot}
\end{table*}

\subsection{Average Verified Confidence as a Proxy for Generation Quality}
On what basis should we dynamically adjust the context? To fully exploit this architectural freedom, we need a reliable, real-time proxy to tell us when the model is struggling. Motivated by recent work on predictive metrics and test-time methods \cite{fu2026deep,kang2025scalable}, we investigated 5 unique predictive metrics derived from DLMs' internal token distributions: (1) Average speculative entropy (ASPE), which is the average entropy at the unmasked indices over the block. (2) Average speculative confidence (ASPC), which is the confidence of speculative tokens at masked indices. (3) Average selected confidence (ASEC): the average confidence of the selected tokens, obtained when the token gets unmasked. (4) Average verified entropy (AVE): the average entropy of the last call over the block, obtained at the last call on the block level. (5) Average verified confidence: the average confidence of the last DLM's call over the selected tokens, obtained at the last call on the block level.

Through a rigorous empirical comparison of correlations between accuracy and the scaling of the number of examples, we identified \textbf{average verified confidence (AVC)} as the most suitable signal, exhibiting the strongest correlation ($r=0.92$) with accuracy across varying numbers of in-context examples (Fig. \ref{fig:N-ACC}). The AVC is defined as:
\begin{equation}
    \bar{C} = \frac{1}{L_{gen}}\sum^{L_{gen}}_{i=1} C_i \;,
\end{equation}
where $L_{gen}$ denotes the length of the generated tokens, and $C_i$ denotes the confidence of the DLM's last call on the $i$-th generated token. Additional comparison results can be found in Appendix \ref{apd: comparison}.

\subsection{Example Insertion Policy} 

Given a pool of ranked examples and our established quality signal (AVC), DIP must choose how many examples to use and update the prompt accordingly. An optimal insertion policy maximizes inference throughput by minimizing the context length, while strictly intervening to preserve generation quality when the model's certainty drops. To address this, we propose an example-insertion policy governed by the dynamic fluctuations of the AVC throughout the generation process.

\noindent\textbf{Confidence Change Ratio.}
We formally define the confidence-change ratio used by our proposed policy within the Fast-dLLM inference framework. Given an average newly generated token confidence for the current $j$-th block $\bar C_{cur}$, where the confidence is obtained by the last DLM call on that $j$-th block, we denote the average cached verified confidence for all previously generated tokens from block 1 to block $j-1$ as $\bar C_{prev}$:
\begin{equation}
    \bar C_{prev} = \frac{1}{j-1}\sum_{k=1}^{j-1} \bar C_k .
\end{equation}We further define the confidence change ratio against previously generated tokens $R$ as:
\begin{equation}
    R(\bar C_{j})  = 1 - \bar C_{cur}/\bar C_{prev},
\end{equation}
where $R<0$ indicates that the newly generated block has higher verified confidence than the previous generation, and $R>0$ indicates a degradation in verified confidence.

\noindent\textbf{Final Policy.} The final policy can be formally defined as follows:
\begin{equation}
\pi(b) \hspace{-1mm} = \hspace{-1mm}
\begin{cases}
    \text{n}_{0} & \text{if } b = 1 \\
    \text{n}_{0} & \text{if } b = 2 \text{ } \& \text{ }  \bar{C}_{1} \ge \text{CST}\\
    \text{n}_{0} + 1 & \text{if } b = 2 \text{ } \& \text{ } \bar{C}_{1} < \text{CST}\\
    \text{n}_{b-1} & \text{if } b \geq 3\text{ } \& \text{ } R(\bar C_{b-1}) < \text{RT} \\
    \text{n}_{b-1} + 1 & \text{if } b \geq  3\text{ } \& \text{ } R(\bar C_{b-1}) \geq \text{RT}
\end{cases}
\end{equation}
where $b$ denotes the current block index, $n_0$ denotes the initial number of examples for block 1, CST stands for cold start confidence threshold for block 2 decision, RT stands for the ratio threshold, and $n_{b-1}$ stands for the number of examples used in the block $(b-1)$.

\begin{table*}[ht]
\centering
\begin{tabular}{|l|l|l|l|l|l|}

\hline
&Dataset&\multicolumn{2}{c|}{GSM8K (5 shot)}  &\multicolumn{2}{|c|}{Math 500 (5 shot)}\\
\hline
Model & Method     & Throughput & Acc & Throughput & Acc \\ \hline
&Fast-dLLM Fixed Prompt            &      75.87(\textcolor{blue}{1.0$\times$})      &    77.7     & 71.69(\textcolor{blue}{1.0$\times$}) & 37.8 \\
LLaDA-8B&DIP($n_0$=1, RT=0.00)    &  \textbf{87.08}(\textcolor{blue}{1.15$\times$})  &    \textbf{77.5}  &75.50(\textcolor{blue}{1.05$\times$}) &35.4\\
\ \ Instruct&DIP($n_0$=2, RT=0.00)&  83.85(\textcolor{blue}{1.11$\times$})  &    77.1    &74.29(\textcolor{blue}{1.04$\times$}) &35.4\\
&DIP($n_0$=1, RT=0.05)           &  89.42(\textcolor{blue}{1.18$\times$})  &    76.8      & 77.15(\textcolor{blue}{1.08$\times$})&35.2\\
&DIP($n_0$=2, RT=0.05)&   86.38(\textcolor{blue}{1.14$\times$})  
&   76.3      & \textbf{75.76}(\textcolor{blue}{1.06$\times$})  &\textbf{37.2}\\

\hline
&Fast-dLLM Fixed Prompt             &      69.94(\textcolor{blue}{1.0$\times$})      &    79.8     &  55.23(\textcolor{blue}{1.0$\times$}) &35.8  \\
\multirow{2}{*}{LLaDA 1.5}&DIP($n_0$=1, RT=0.00)   &  86.55(\textcolor{blue}{1.24$\times$})  &   78.9     &76.26(\textcolor{blue}{1.38$\times$}) &34.4\\
&DIP($n_0$=2, RT=0.00)&  82.57(\textcolor{blue}{1.18$\times$})  &   78.7     & 74.02(\textcolor{blue}{1.34$\times$})&37.2\\
&DIP($n_0$=1, RT=0.05)           &   89.18(\textcolor{blue}{1.28$\times$})  &   78.8   & 77.91(\textcolor{blue}{1.41$\times$})&34.8\\
&DIP($n_0$=2, RT=0.05)&    \textbf{85.63}(\textcolor{blue}{1.22$\times$})  &     \textbf{79.5}  & \textbf{75.69}(\textcolor{blue}{1.37$\times$})&\textbf{38.4}\\ \hline
\end{tabular}
\caption{DIP's results on 5-shot GSM8K and 5-shot MATH500 compared  against fixed prompt baseline using LLaDA-8B-Instruct \citep{nie2025large} and LLaDA-1.5 \cite{zhu2025llada}.}
\label{tab:main-5-shot}
\end{table*}

\section{Experiment} 
\subsection{Experiment Setup}
\noindent\textbf{Dataset and Tasks.}
Following previous studies \cite{wu2025fast} and considering the availability of few-shot examples, we evaluate DIP under 5-shot and 8-shot settings on three generative tasks: (1) GSM8K \cite{cobbe2021training} consists of math word problems, (2) MATH500 \cite{hendrycksmath2021,lewkowycz2022solving}, a high-quality subset of math problems spanning 7 topics from high school math competitions, and (3) MBPP \cite{austin2021program}, a coding benchmark consists of Python programming problems.
To ensure reproducibility and the use of identical few-shot (in-context) examples under the n-shot setting, our experiment uses the \textit{lm-eval} library \citep{eval-harness}. 
\begin{table}[]
\centering
\begin{tabular}{|l|l|l|}
\hline
Ranking Method & Throughput     & Acc \\
\hline
Random Order &84.06&77.9 \\
MMR($\lambda=0.2$) &83.88& 78.4 \\
MMR($\lambda=0.4$) &84.12& 78.2\\
MMR($\lambda=0.6$) &83.93& 77.2\\
MMR($\lambda=0.8$) &83.90& 77.9\\
Cosine Similarity &84.26& 78.6\\
\hline
\end{tabular}
\caption{Ablation study of DIP's different ranking methods on 8-shot GSM8K with LLaDA-8B-Instruct \citep{nie2025large}. }
\label{tab:aba-ranking}
\end{table}

\noindent\textbf{Evaluation Metrics.} We use two metrics: (1) inference throughput measured by the average number of output tokens generated over total runtime (including embedding, ranking, and prompt update latency), and (2) accuracy.

\noindent\textbf{Baseline.} We compare DIP against \textit{Fast-dLLM Fixed Prompt}, a strong, state-of-the-art baseline representing standard KV cache-enhanced inference with a static context. We evaluate both approaches under 5- and 8-shot settings \cite{wu2025fast}. We sample the in-context examples with the \texttt{lm-eval} package and present them in random order. All results were obtained using the official Fast-dLLM codebase with the default configurations: a generation length of 256 tokens, a block size of 32, and 8 blocks (unless otherwise noted).

\noindent\textbf{Models and Implementation Details.}
Following prior work \cite{li2026beyond,jin2025thinking}, we experiment with the baseline and DIP on two representative masked DLMs: LLaDA-8B-Instruct \cite{nie2025large} and LLaDA-1.5 \cite{zhu2025llada}. To balance computational efficiency and sentence-embedding quality, we use Qwen3-Embedding-0.6B \cite{zhang2025qwen3, reimers-gurevych-2019-sentence} for obtaining the embeddings of questions. We run all experiments on NVIDIA RTX 5090 GPUs to ensure a fair comparison (unless otherwise noted).

\subsection{Main Results}
In Tables \ref{tab:main-8-shot} and \ref{tab:main-5-shot}, we compare DIP's performance against the fixed-prompt baseline, using cosine similarity and $CST$=0.93 (unless otherwise noted) in the 5-shot and 8-shot settings on two math reasoning tasks (MATH500 and GSM8K). In summary, the results show that DIP consistently improves inference throughput while largely preserving, and in some cases even improving, the generation quality of the fixed-prompt baseline. 

As shown in Table \ref{tab:main-8-shot} in the 8-shot setting, DIP notably achieves up to 1.50$\times$ inference speedup on the GSM8K task and 1.59$\times$ on MATH500 with LLaDA-1.5 while largely preserving the generation quality. Results also show improved inference throughput on LLaDA-8B-Instruct, with up to 1.36$\times$ on GSM8K and 1.23$\times$ on the MATH500 dataset. 
In the 5-shot setting (Table \ref{tab:main-5-shot}), we observed that DIP's inference still improves inference speed, with throughput similar to that of 8 shots. This suggests that DIP can benefit from additional examples without incurring too much computational cost as the number of examples increases. 

For the coding task, namely MBPP, experiments show a similar pattern in DIP's effectiveness, yielding up to a 1.45$\times$ speedup with the LLaDA-8B-Instruct Model. Due to limited space, MBPP results are included in the Appendix \ref{apd:mbpp}. We also include additional stability analysis in the Appendix \ref{apd:variance}.

\subsection{Ablation Study and Additional Analysis}

\noindent\textbf{Ranking Methods.} To further understand the example order's influence on DIP, we conducted an ablation over the ranking methods: (1) cosine similarity and (2) MMR with different $\lambda$. As shown in Table \ref{tab:aba-ranking}, the order of example insertion does not influence the throughput, while improving accuracy with the right ordering. More results with the MMR example ranking are in Appendix \ref{apd:results}. 

\begin{figure}[t]
\centering
  \includegraphics[width=\columnwidth]{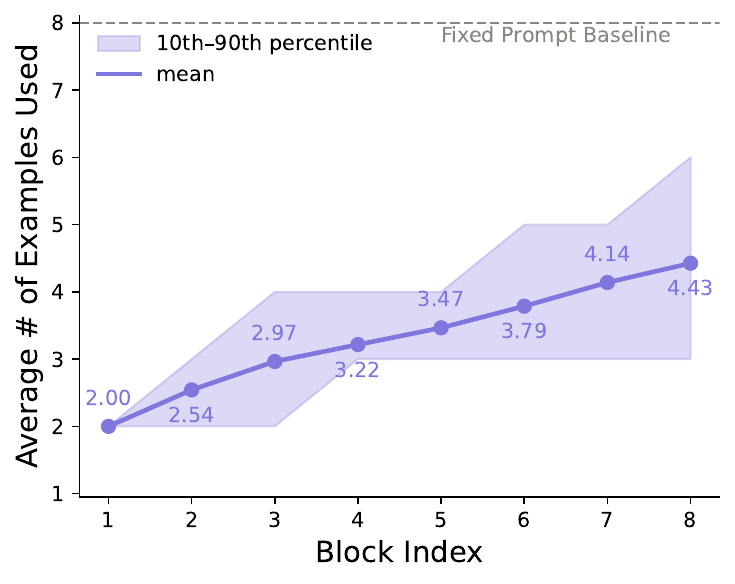}
  \caption{\label{fig:EU-math} Example usage distribution along the block index on 8-shot GSM8K with LLaDA-8B-Instruct.}
\end{figure}

\begin{figure}[t]
\centering
  \includegraphics[width=\columnwidth]{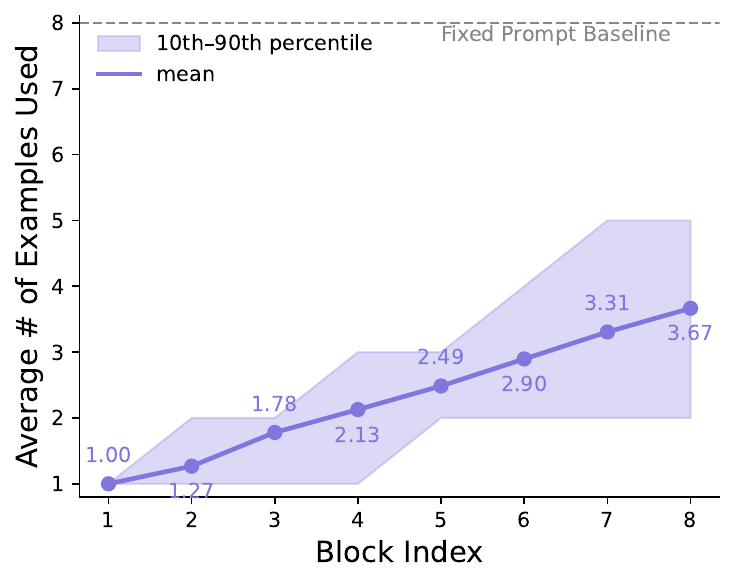}
  \caption{\label{fig:EU-MATH500} Example usage distribution along the block index on 8-shot MATH500 with LLaDA-1.5.}
\end{figure}

\noindent\textbf{Example Usage By Block and Speedup Analysis.}
Fig. \ref{fig:EU-math} presents the example usage distribution by block index on 8-shot GSM8K with the LLaDA-8B-Instruct model. We obtained this data using DIP with $n_0=2$, $RT=0$, $CST=0.93$, and cosine-similarity ranking, yielding a 1.26$\times$ speedup over the fixed-prompt baseline. 
In Fig. \ref{fig:EU-MATH500}, results show similar trends in 8-shot MATH500 with the LLaDA-1.5 model. This data was obtained using DIP with $n_0=1$, $RT=0$, $CST=0.93$, and cosine-similarity ranking, yielding a 1.56$\times$ speedup over the fixed-prompt baseline. 
As shown, DIP's example-insertion policy yields greater speedup in the early stages than in the later stages, contributing to higher throughput.

\noindent\textbf{Hyperparameter Sensitivity.}
To further investigate DIP's hyperparameter sensitivity, we conduct an ablation study to evaluate the impact of hyperparameter choices, using LLaDA-1.5 on the 8-shot GSM8K. The results in Appendix \ref{apd:hyper} provide clear guidance for selecting hyperparameters: larger $n_0$, larger $RT$, and larger $CST$ are preferred for accuracy. Smaller $n_0$, larger $RT$, and smaller $CST$ are preferred to maximize inference throughput.

\noindent\textbf{Performance Across 256 and 512 Generation Lengths.}
As noted in Eq. \ref{eq:reduction}, the overall computation expense is based on two factors: (1) the length of the prompt and (2) generation lengths. Increasing generation length could reduce speedup. To further investigate performance differences across generation lengths, we compared LLaDA-8B-Instruct and LLaDA-1.5 in the 8-shot GSM8K setting with generation lengths of 256 and 512. As shown in Table \ref{tab:aba-length}, consistent with Eq. \ref{eq:reduction}, the speedup decreases as the generation length increases from 256 to 512. Overall, DIP can still yield up to a 1.21$\times$ speedup at 512-token generation length and 1.42$\times$ speedup at 256-token generation length. Due to memory constraint, this experiment is performed using NVIDIA RTX PRO 6000 GPUs.




\begin{table*}[ht!]
\centering
\begin{tabular}{|l|l|l|l|l|}
\hline
Model & Method & Gen Len & Throughput     & Acc \\
\hline
\multirow{4}{*}{LLaDA-8B-Instruct}
&Fast-dLLM Fixed Prompt& 256  &66.58(\textcolor{blue}{1.0$\times$}) & 77.3\\
&DIP($n_0=2,RT=0)$&256&84.08(\textcolor{blue}{1.26$\times$})&78.6 \\ \cline{2-5}
&Fast-dLLM Fixed Prompt& 512  & 43.18(\textcolor{blue}{1.0$\times$})&74.7\\
&DIP($n_0=2,RT=0)$&512&52.46(\textcolor{blue}{1.21$\times$})& 75.6 \\
\hline
\multirow{4}{*}{LLaDA 1.5}&Fast-dLLM Fixed Prompt& 256  &58.60(\textcolor{blue}{1.0$\times$}) & 80.1\\
&DIP($n_0=2,RT=0)$&256&82.92(\textcolor{blue}{1.42$\times$})&80.3 \\
\cline{2-5}
&Fast-dLLM Fixed Prompt& 512  &44.70(\textcolor{blue}{1.0$\times$}) &78.2\\
&DIP($n_0=2,RT=0)$&512&53.10(\textcolor{blue}{1.19$\times$})& 79.2 \\

\hline
\end{tabular}
\caption{Performance comparison between 256 and 512 under the 8-shot GSM8K Setting. }
\label{tab:aba-length}
\end{table*}

\noindent\textbf{Prompt Change In KV-Cache Enhanced DLMs Inference Framework.}
Because DIP aims to optimize inference throughput in DLMs, it must be compatible with the KV cache-enhanced inference framework. DIP is based on Fast-dLLM \citep{wu2025fast}, a KV cache-enhanced block-wise inference framework, which requires a necessary non-KV cache call at the start of discrete block decoding to update the KV cache (Appendix \ref{apd:algo}). 

In principle, DIP is a plug-in method that works with any masked DLM KV-cache-enhanced inference framework, and a prompt update can occur alongside any non-KV-cache call.


\section{Conclusion}

In this paper, we make a key discovery: DLMs enable \textit{low-cost dynamic context adjustment}. We also empirically evaluate correlations between generation accuracy and token confidence for DLMs and identify average verified confidence as the key signal for guiding context adjustment. 
Building on these insights, we propose DIP, a practical context optimization framework that dynamically scales in-context examples during generation based on verified confidence.
Experimental results on math and coding benchmarks with LLaDA-1.5 and LLaDA-8B-Instruct show that DIP achieves up to $1.59\times$ and $1.36\times$ speedups, respectively, while largely preserving the generation quality of the fixed-prompt baseline. Our key findings open a new avenue for efficient DLM inference through dynamic context optimization.

\section*{Acknowledgments}
This research was supported by the National Science Foundation under IIS-2348405, 2442253, 2607580, NIH 1R21AG091260-01, Commonwealth Cyber Initiative, and generous gifts from Nvidia, Cisco, and the Amazon-Virginia Tech Initiative. 
The authors acknowledge the National Artificial Intelligence Research Resource (NAIRR) Pilot (through allocation NAIRR250123 and NAIRR240202), William \& Mary Research Computing, UIUC Delta, and Purdue Anvil supercomputers for contributing to this research result.

\section*{Limitations}
Our proposed approach has a few limitations. First, the training-free nature of DIP introduces a trade-off: the example-usage policy depends on hyperparameters that must be empirically tuned to accommodate varying task difficulties. Second, we constrain our exploration of DLMs' architectural freedom strictly to example usage in in-context learning. While dynamic context in DLMs holds significant potential for applications beyond ICL, we leave these broader structural investigations to future work.

\bibliography{custom}

\clearpage

\appendix


\section{Appendix: An Example Context Change Through Different Blocks Using MATH500} \label{apd:example}

Context for block 1 with $n_0=1$, i.e., one example:
\begin{lstlisting}
Problem:
Find the domain of the expression  $\frac{\sqrt{x-2}}{\sqrt{5-x}}$.}

Solution: The expressions inside each square root must be non-negative. Therefore, $x-2 \ge 0$, so $x\ge2$, and $5 - x \ge 0$, so $x \le 5$. Also, the denominator cannot be equal to zero, so $5-x>0$, which gives $x<5$. Therefore, the domain of the expression is $\boxed{[2,5)}$.
Final Answer: The final answer is $[2,5)$. I hope it is correct.

Problem:
Define
\[p = \sum_{k = 1}^\infty \frac{1}{k^2} \quad \text{and} \quad q = \sum_{k = 1}^\infty \frac{1}{k^3}.\]Find a way to write
\[\sum_{j = 1}^\infty \sum_{k = 1}^\infty \frac{1}{(j + k)^3}\]in terms of $p$ and $q.$

Solution:
\end{lstlisting}

\noindent Context for block 2 and block 3 with two examples:
\begin{lstlisting}
Problem:
Find the domain of the expression  $\frac{\sqrt{x-2}}{\sqrt{5-x}}$.}

Solution: The expressions inside each square root must be non-negative. Therefore, $x-2 \ge 0$, so $x\ge2$, and $5 - x \ge 0$, so $x \le 5$. Also, the denominator cannot be equal to zero, so $5-x>0$, which gives $x<5$. Therefore, the domain of the expression is $\boxed{[2,5)}$.
Final Answer: The final answer is $[2,5)$. I hope it is correct.

Problem:
If $\det \mathbf{A} = 2$ and $\det \mathbf{B} = 12,$ then find $\det (\mathbf{A} \mathbf{B}).$

Solution: We have that $\det (\mathbf{A} \mathbf{B}) = (\det \mathbf{A})(\det \mathbf{B}) = (2)(12) = \boxed{24}.$
Final Answer: The final answer is $24$. I hope it is correct.

Problem:
Define
\[p = \sum_{k = 1}^\infty \frac{1}{k^2} \quad \text{and} \quad q = \sum_{k = 1}^\infty \frac{1}{k^3}.\]Find a way to write
\[\sum_{j = 1}^\infty \sum_{k = 1}^\infty \frac{1}{(j + k)^3}\]in terms of $p$ and $q.$

Solution:

+ GENERATED CONTENT.
\end{lstlisting}

\noindent Context for block 4 with three examples:
\begin{lstlisting}
Problem:
Find the domain of the expression  $\frac{\sqrt{x-2}}{\sqrt{5-x}}$.}

Solution: The expressions inside each square root must be non-negative. Therefore, $x-2 \ge 0$, so $x\ge2$, and $5 - x \ge 0$, so $x \le 5$. Also, the denominator cannot be equal to zero, so $5-x>0$, which gives $x<5$. Therefore, the domain of the expression is $\boxed{[2,5)}$.
Final Answer: The final answer is $[2,5)$. I hope it is correct.

Problem:
If $\det \mathbf{A} = 2$ and $\det \mathbf{B} = 12,$ then find $\det (\mathbf{A} \mathbf{B}).$

Solution: We have that $\det (\mathbf{A} \mathbf{B}) = (\det \mathbf{A})(\det \mathbf{B}) = (2)(12) = \boxed{24}.$
Final Answer: The final answer is $24$. I hope it is correct.

Problem:
Terrell usually lifts two 20-pound weights 12 times. If he uses two 15-pound weights instead, how many times must Terrell lift them in order to lift the same total weight?

Solution: If Terrell lifts two 20-pound weights 12 times, he lifts a total of $2\cdot 12\cdot20=480$ pounds of weight.  If he lifts two 15-pound weights instead for $n$ times, he will lift a total of $2\cdot15\cdot n=30n$ pounds of weight.  Equating this to 480 pounds, we can solve for $n$:
\begin{align*}
30n&=480\
\Rightarrow\qquad n&=480/30=\boxed{16}
\end{align*}
Final Answer: The final answer is $16$. I hope it is correct.

Problem:
Define
\[p = \sum_{k = 1}^\infty \frac{1}{k^2} \quad \text{and} \quad q = \sum_{k = 1}^\infty \frac{1}{k^3}.\]Find a way to write
\[\sum_{j = 1}^\infty \sum_{k = 1}^\infty \frac{1}{(j + k)^3}\]in terms of $p$ and $q.$

Solution: 

+ GENERATED CONTENT.
\end{lstlisting}

\section{Appendix: Algorithm for Fast-dLLM and DIP} \label{apd:algo}
Algorithm \ref{alg:fast-dllm} presents the decoding process of Fast-dLLM \cite{wu2025fast}. Building on Fast-dLLM, we present the algorithm of DIP in Algorithm \ref{alg:dip}. 
\begin{algorithm*}[t]
\caption{Fast-dLLM Decoding with Threshold}
\label{alg:fast-dllm}
\KwIn{$\text{model }p_\theta $, prompt $p$, answer length $L$, blocks $N$, block size $B$, step per block $T$, threshold $\tau$}

$x \leftarrow$ [$p$;\text{[MASK],...,[MASK]}] 

Initialize KV Cache for $x$. \hfill $\rhd\!\rhd\!\rhd\!$ KV Cache Init

\For{$n=$1 to $N$}{
    $s \leftarrow$ [$| p |$ + $(n-1)B$], $e = |p| + nB$
    
    \textbf{Update KV Cache with non-KV cache call} \hfill $\rhd\!\rhd\!\rhd\!$ KV Cache Update
    
    \For{$t=1$ \text{to} $T$}{
    Use cache, call $p_\theta$ on $x^{[s,e)]}$ \hfill $\rhd\!\rhd\!\rhd\!$ KV Cache Reuse
    
    For each masked $x^i$, confidence $c^i = \text{max}_x p_\theta(x^i)$   

    Unmask all $i$ in $[s,e)$ with $c^i \ge \tau$, with index max $c^i$
    
    \If{all $x^{[s,e)} != \text{[MASK]}$}{
    \textbf{break, move to next block}
    }

    }

}
\Return$x$;
\end{algorithm*}

\begin{algorithm*}[h]
\caption{DIP with Fast-dLLM}
\label{alg:dip}
\KwIn{$\text{model }p_\theta $, answer length $L$, blocks $N$, block size $B$, step per block $T$, threshold $\tau$, example insertion policy $\pi$, examples set $E$, number of examples $K$, query $q$}

Rank $E$ using defined ranking method, $E_{rank}$ = $[E_1,E_2,...,E_K]$ \hfill $\rhd\!\rhd\!\rhd\!$ \textcolor{blue}{Example Ranking Stage}

$p \leftarrow$ [$E_1;q$], $k = 1$ \hfill $\rhd\!\rhd\!\rhd\!$ \textcolor{red}{Initialize Prompt With $\pi$}

$x \leftarrow$ [$p ; \text{[MASK],...,[MASK]}$] 

Initialize KV Cache for $x$. \hfill $\rhd\!\rhd\!\rhd\!$ KV Cache Init

$\bar\mu = 0$ \hfill $\rhd\!\rhd\!\rhd\!$ Average Verified Confidence Over All Generated Tokens

$\mu = 0$ \hfill $\rhd\!\rhd\!\rhd\!$ Verified Confidence Mean Of Current Block

\For{$n=$1 to $N$}{
    $s \leftarrow$ [$| p |$ + $(n-1)B$], $e = |p| + nB$
    
    \If{$n!=1$}{      

        Insertion = $\pi(\mu, \bar\mu)$  \hfill $\rhd\!\rhd\!\rhd\!$ \textcolor{red}{Get Action Following $\pi$}
    
        \If{$\text{Insertion} == True$}{ 
        
            $p \leftarrow$ \text{[}$E_1,E_2,..,E_k;q$], $k += 1$ \hfill $\rhd\!\rhd\!\rhd\!$ \textcolor{red}{Update Prompt}

            $lp = $ length of $p$

            $x \leftarrow$ [$p ; \text{[MASK],...,[MASK]}$] 
        }

        \textbf{Update KV Cache with non-KV cache call} \hfill $\rhd\!\rhd\!\rhd\!$ KV Cache Update
    }

    \For{$t=1$ \text{to} $T$}{
    Use cache, call $p_\theta$ on $x^{[s,e)]}$ \hfill $\rhd\!\rhd\!\rhd\!$ KV Cache Reuse
    
    For each masked $x^i$, confidence $c^i = \text{max}_x p_\theta(x^i)$   

    Unmask all $i$ in $[s,e)$ with $c^i \ge \tau$, with index max $c^i$
    
    \If{all $x^{[s,e)} != \text{[MASK]}$}{
    $\mu$ = mean($c^i$ of $x^{[s,e)}$) \hfill $\rhd\!\rhd\!\rhd\!$ Save Confidence Mean Of Current Block
    
    $\bar\mu$ = mean($c^i$ of $x^{[lp,e)}$) \hfill $\rhd\!\rhd\!\rhd\!$ Save Confidence Mean Of Generated Tokens
    
    \textbf{break, move to next block}
    }

    $\mu$ = mean($c^i$ of $x^{[s,e)}$) \hfill $\rhd\!\rhd\!\rhd\!$ Save Confidence Mean Of Current Block
    
    $\bar\mu$ = mean($c^i$ of $x^{[lp,e)}$) \hfill $\rhd\!\rhd\!\rhd\!$ Save Confidence Mean Of Generated Tokens
    
    }
}
\Return$x$;
\end{algorithm*}

\section{Appendix: Additional Metrics Comparison}\label{apd: comparison}
In this section, we define metrics specific to DLMs' generation process and present the impact of the number of in-context examples on metrics and accuracy. We collect all data from the same 1000 sampled questions in the GSM8K training set, with 0 to 8 in-context examples.

\begin{figure}[h]
\centering
  \includegraphics[width=\columnwidth]{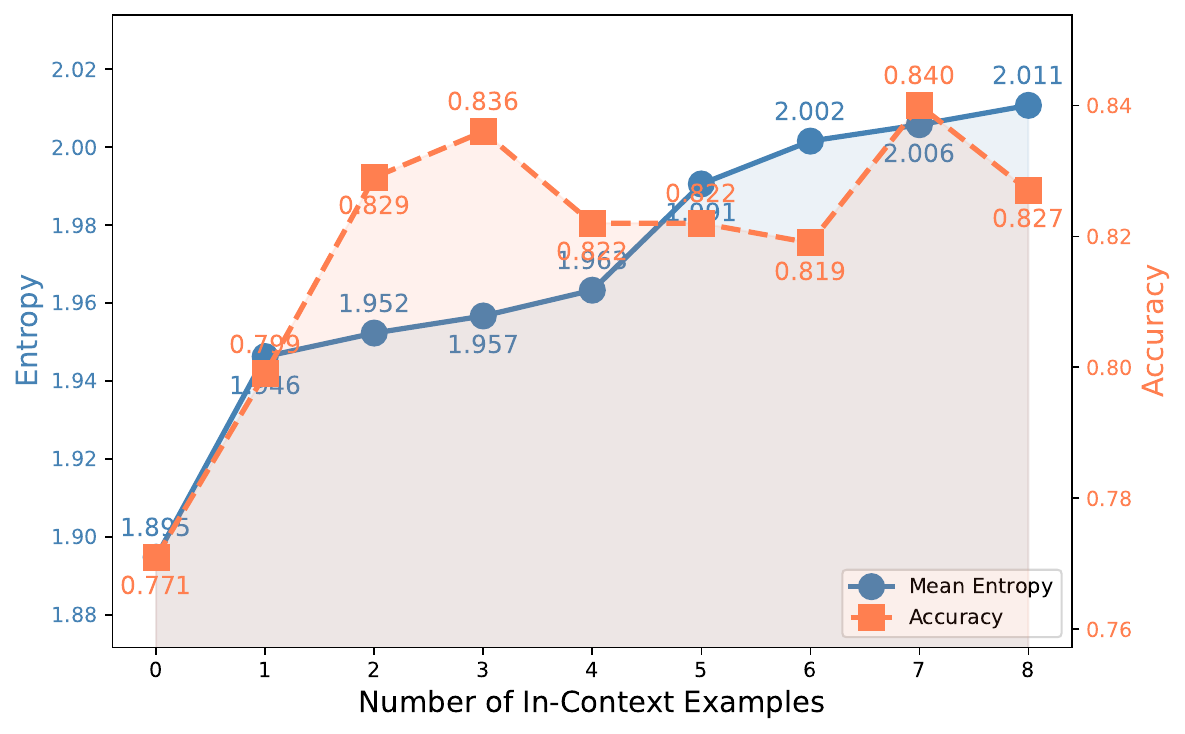}
  \caption{\label{fig:CA-spec-entro} Impact of number of in-context examples vs. average speculative entropy and accuracy.}
\end{figure}

\textbf{(1) Speculative Entropy}: the entropy at the unmasked index, which is obtained at the Non-KV cache call. In Fig. \ref{fig:CA-spec-entro}, this metric shows a positive correlation ($r=+0.76$) with accuracy.

\begin{figure}[h]
\centering
  \includegraphics[width=\columnwidth]{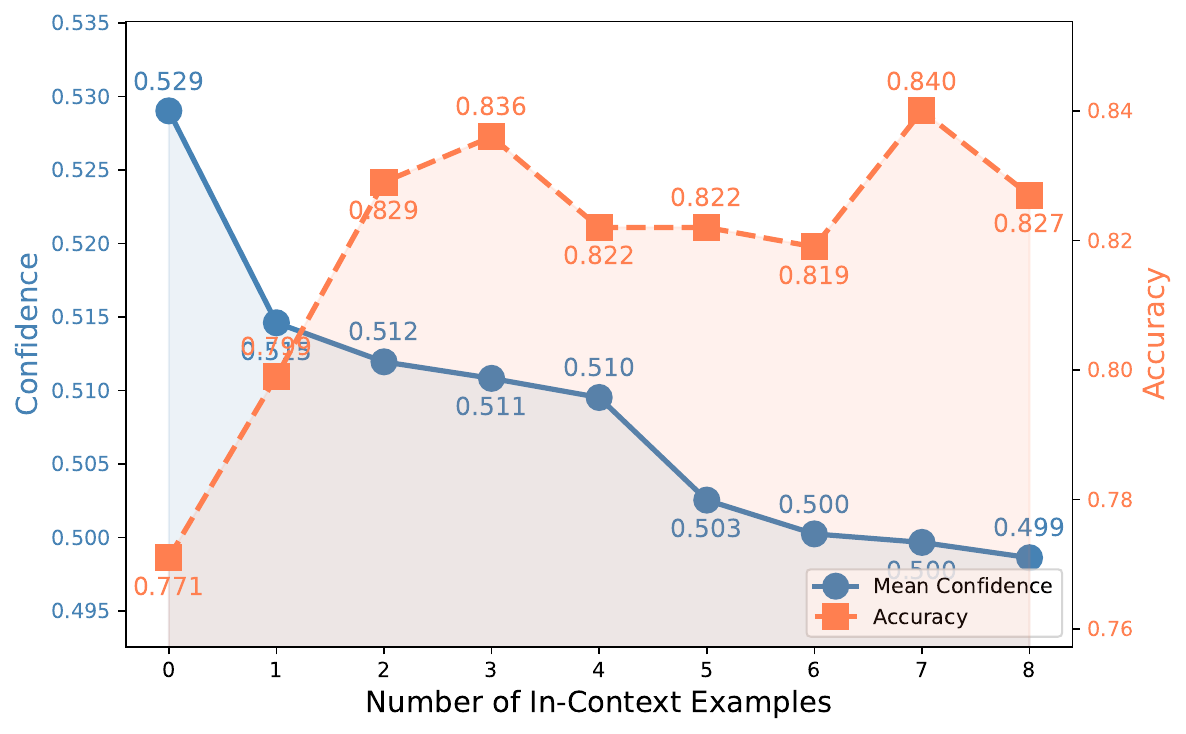}
  \caption{\label{fig:CA-spec-conf} Impact of number of in-context examples vs. average speculative confidence and accuracy. }
\end{figure}

\textbf{(2) Speculative Confidence}: the highest confidence at a masked index, which is obtained at the Non-KV cache call. In Fig. \ref{fig:CA-spec-conf}, this metric shows a negative correlation ($r=-0.79$) with accuracy. 

\begin{figure}[h]
\centering
  \includegraphics[width=\columnwidth]{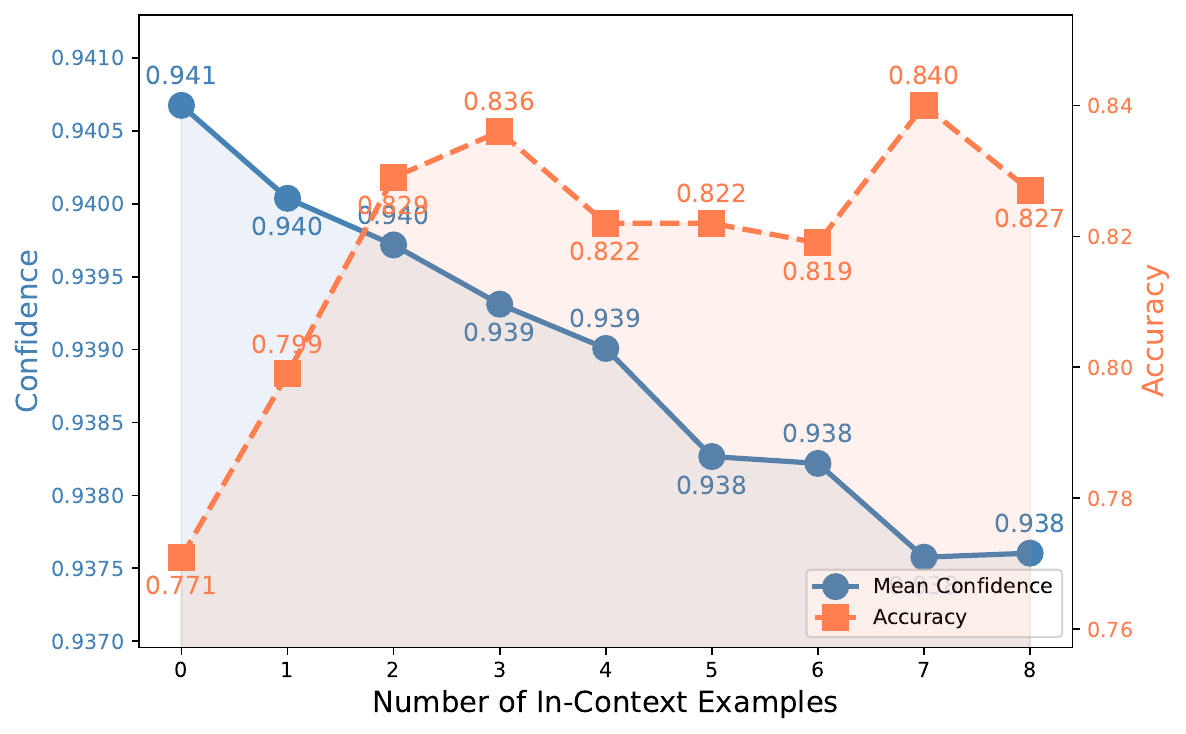}
  \caption{\label{fig:CA-sele-conf} Impact of number of in-context examples vs. average selected confidence and accuracy. }
\end{figure}

\textbf{(3) Selected Confidence}: the confidence of the selected token, which is obtained when the token gets unmasked. This metric is similar to token confidence in ARLMs \cite{fu2026deep}. In Fig. \ref{fig:CA-sele-conf}, this metric shows a negative correlation ($r= -0.71$) with accuracy. 
\begin{figure}[h]
\centering
  \includegraphics[width=\columnwidth]{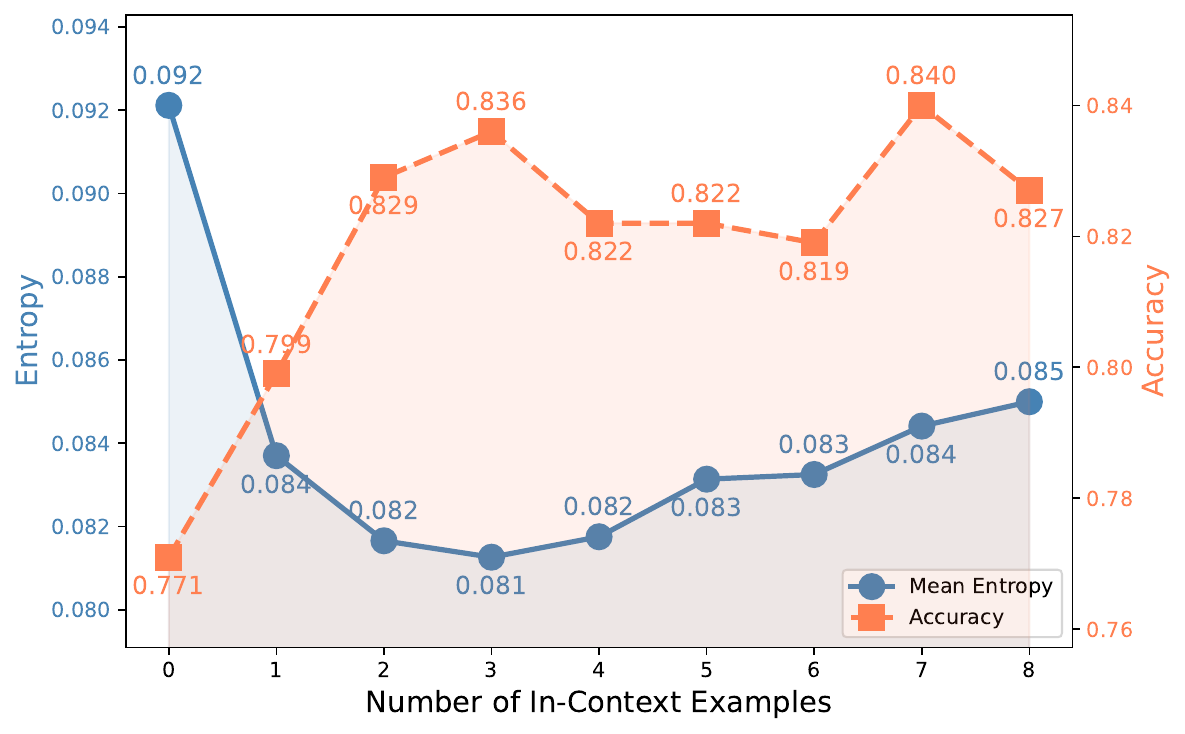}
  \caption{\label{fig:CA-verifi-entropy} Impact of number of in-context examples vs. average verified entropy and accuracy. }
\end{figure}

\textbf{(4) Verified Entropy}: the entropy of the last call over the index, which is usually the last call on the block level in Fast-dLLM. In Fig. \ref{fig:CA-verifi-entropy}, this metric shows a negative correlation ($r=-0.80$) with accuracy.

\textbf{(5) Verified Confidence}: the verified confidence of the selected token at the last call over the token, which is usually the last call on the block level in Fast-dLLM. In Fig. \ref{fig:N-ACC}, this metric shows a positive correlation ($r=0.92)$ with accuracy.

\noindent\textbf{Summary}: By comparing those relationships from different metrics, average verified confidence  (Fig. \ref{fig:Conf-Corr} ) shows the strongest correlation with accuracy among all those predictive metrics derived from DLMs' internal token distributions. Hence, we used the average verified confidence metric as the quality signal throughout our experiments.

\section{Appendix: MBPP Results}\label{apd:mbpp}
We evaluate DIP on the MBPP benchmark using both the LLaDA-8B-Instruct and LLaDA-1.5 models. As detailed in Tables \ref{tab:mbpp-table-8bi} and \ref{tab:mbpp-table-1.5}, DIP consistently accelerates inference, achieving speedups of up to $1.45\times$ while maintaining comparable generation quality. Notably, because code generation tasks inherently demand rigorous logical and syntactic reasoning, empirical results indicate that DIP requires a larger initial number of in-context examples to effectively bootstrap the generation process. 

\section{Appendix: Stability Results}\label{apd:variance}
To further investigate DIP stability and account for the inherent noise in DLM generation, we conducted multi-run evaluations in two representative settings.

First, we evaluated the configuration that showed the most significant drop in accuracy in our main results: LLaDA-8B-Instruct in the MATH500 8-shot setting. Across 5 independent runs, the Fast-dLLM fixed-prompt baseline achieved an accuracy of 37.5 $\pm$ 1.34, while DIP ($n_0=1$, $RT=0.00$) achieved 36.0 $\pm$ 2.21. Although DIP shows higher variance in this boundary-case configuration, its overall performance remains competitive. 

Second, we evaluated our optimal configuration on GSM8K with LLaDA-1.5 using DIP ($n_0=2$, $RT=0.00$) across 3 independent runs for both the 5-shot and 8-shot settings. As detailed in Table \ref{tab:apd-variance}, the average accuracy and throughput are highly stable and consistent with the single-run values reported in our main results, demonstrating that DIP reliably maintains generation quality while preserving its speedup advantages. 

\begin{table}[ht]
\centering
\begin{tabular}{|l| l| l| l|}
\hline
Shot & Method & Throughput & Accuracy \\
\hline
5 & Baseline & 72.52 $\pm$ 0.9 & 80.1 $\pm$ 0.9 \\
5 & DIP & 88.78 $\pm$ 0.7 & 79.9 $\pm$ 1.5 \\
\hline
8 & Baseline & 61.96 $\pm$ 0.3 & 80.6 $\pm$ 0.5 \\
8 & DIP & 85.87 $\pm$ 0.9 & 80.2 $\pm$ 0.4 \\
\hline
\end{tabular}
\caption{Results with LLaDA-1.5 on 5 and 8-shot GSM8K across 3 independent runs. Values denote Mean $\pm$ Standard Deviation.}
\label{tab:apd-variance}
\end{table}

\section{Appendix: Additional MMR Results}\label{apd:results}
In this section, we present additional results with the MMR example ranking method. Following the results in Table \ref{tab:aba-ranking}, we experiment with MMR example ranking with $\lambda=0.2$. In Table \ref{tab:apd-mmr1}, we present results on 8-shot GSM8K using the LLaDA-8B-Instruct model alone, with the fixed prompt baseline. In Table \ref{tab:apd-mmr2}, we present results on the 8-shot GSM8K task using the LLaDA-1.5 model alone, with the fixed prompt baseline. 
\begin{table}[ht]
\centering
\begin{tabular}{|l|l|l|}
\hline
DIP Setting & Throughput     & Acc \\
\hline
Baseline&66.58&77.3\\
$n_0$=1, RT=0.00 &87.13&75.4 \\
$n_0$=2, RT=0.00 &83.98&78.4  \\
$n_0$=1, RT=0.05 &89.89&76.0 \\
$n_0$=2, RT=0.05 &86.91&77.3 \\
\hline
\end{tabular}
\caption{Additional results on 8-shot GSM8K with MMR example ranking where $\lambda=0.2$ on LLaDA-8B-Instruct.}
\label{tab:apd-mmr1}
\end{table}

\section{Appendix: Hyperparameter Sensitivity Results}\label{apd:hyper}

We present the sensitivity results over DIP's hyperparameter choices on 8-shot GSM8K using LLaDA-1.5 \cite{zhu2025llada} in Table \ref{tab:aba-N0}, Table \ref{tab:aba-rt}, and Table \ref{tab:aba-CST}. Results show that a wide range of parameters can yield reasonable performance compared to the Fast-dLLM Fixed Prompt with speedups on the GSM8K task.



\begin{table*}[ht]
\centering
\begin{tabular}{|l|l|l|l|l|}

\hline
&Dataset& &   \multicolumn{2}{c|}{MBPP}  \\
\hline
Model & Method      & \# of Shot &  Throughput& Acc \\ \hline
LLaDA-8B&Baseline                  & 5        &  41.70(\textcolor{blue}{1.0$\times$}) & 26.6  \\
\ \ Instruct&DIP($n_0$=3, RT=0.00)      &  5  &49.98(\textcolor{blue}{1.20$\times$})&27.6\\
&DIP($n_0$=3, RT=0.05)  &  5 &\textbf{52.37}(\textcolor{blue}{1.25$\times$}) & \textbf{27.4}\\

\hline
LLaDA-8B&Baseline                  & 8        &  34.27(\textcolor{blue}{1.0$\times$}) & 28.0  \\
\ \ Instruct&DIP($n_0$=3, RT=0.00)      &  8  &47.80(\textcolor{blue}{1.39$\times$})&26.4\\
&DIP($n_0$=3, RT=0.05)  &  8 &\textbf{49.61}(\textcolor{blue}{1.45$\times$}) & \textbf{27.8}\\

\hline
\end{tabular}
\caption{DIP's results on MBPP comparing against fixed prompt baseline using LLaDA-8B-Instruct \citep{nie2025large}. \textcolor{blue}{Blue: relative speedup.} We also \textbf{bold} the DIP configuration that achieves the highest speedup while maintaining on-par or higher accuracy. }
\label{tab:mbpp-table-8bi}
\end{table*}

\begin{table*}[ht]
\centering
\begin{tabular}{|l|l|l|l|l|l|}

\hline
&Dataset& &   \multicolumn{2}{c|}{MBPP}  \\
\hline
Model & Method      & \# of Shot &  Throughput& Acc \\ \hline
\multirow{3}{*}{LLaDA 1.5}&Baseline                  & 5        &  40.70(\textcolor{blue}{1.0$\times$}) & 26.0  \\
&DIP($n_0$=3, RT=0.00)      &  5  &\textbf{48.78}(\textcolor{blue}{1.20$\times$})&\textbf{25.0}\\
&DIP($n_0$=3, RT=0.05)  &  5 &50.02(\textcolor{blue}{1.23$\times$}) &23.8\\

\hline
\multirow{3}{*}{LLaDA 1.5 \tablefootnote{Random ordered examples}}&Baseline                  & 8        &  35.29(\textcolor{blue}{1.0$\times$}) & 29.4  \\
&DIP($n_0$=4, RT=0.00)&  8  &\textbf{42.98}(\textcolor{blue}{1.22$\times$})&\textbf{27.8}\\
&DIP($n_0$=4, RT=0.05)  &  8 &44.10(\textcolor{blue}{1.25$\times$}) &26.2\\

\hline
\end{tabular}
\caption{DIP's results on MBPP comparing against fixed prompt baseline using LLaDA-1.5 \cite{zhu2025llada}. \textcolor{blue}{Blue: relative speedup.} We also \textbf{bold} the DIP configuration that achieves the highest speedup while maintaining on-par or higher accuracy. }
\label{tab:mbpp-table-1.5}
\end{table*}

\begin{table}[ht!]
\centering
\begin{tabular}{|l|l|l|}
\hline
 $n_0$ & Throughput     & Acc \\
\hline
Baseline & 58.60  & 80.1\\
$1$ &90.65&78.9 \\
$2$ &85.86&80.3 \\
$3$ &80.82&81.5 \\
$4$ &77.46&80.2 \\
$5$ &74.00&79.4 \\
\hline
\end{tabular}
\caption{Ablation study of initial examples $n_0$ on 8-shot GSM8K with LLaDA-1.5 \citep{zhu2025llada}. }
\label{tab:aba-N0}
\end{table}

\begin{table}[]
\centering
\begin{tabular}{|l|l|l|}
\hline
DIP Setting & Throughput     & Acc \\
\hline
Baseline&58.60&80.1\\
$n_0$=1, RT=0.00 &85.56&79.2\\
$n_0$=2, RT=0.00 &81.20&79.2  \\
$n_0$=1, RT=0.05 &89.68& 79.6\\
$n_0$=2, RT=0.05 &85.80& 79.6\\

\hline
\end{tabular}
\caption{Additional results on 8-shot GSM8K with MMR example ranking where $\lambda=0.2$ on LLaDA-1.5.}
\label{tab:apd-mmr2}
\end{table}

\begin{table}[ht!]
\centering
\begin{tabular}{|l|l|l|}
\hline
 $RT$ & Throughput     & Acc \\
\hline
Baseline & 58.60  & 80.1\\
$-0.1$  &79.11&79.7 \\
$-0.05$ &81.35&79.4 \\
$0$     &86.56&80.3 \\
$0.05$  &89.28&80.2 \\
$0.01$  &88.90&80.5 \\
\hline
\end{tabular}
\caption{Ablation study of ratio threshold $RT$ on 8-shot GSM8K with LLaDA-1.5 \citep{zhu2025llada}. }
\label{tab:aba-rt}
\end{table}

\begin{table}[ht!]
\centering
\begin{tabular}{|l|l|l|}
\hline
 $CST$ & Throughput     & Acc \\
\hline
Baseline & 58.60  & 80.1\\
0.75&87.36&78.9\\
0.8&88.62&78.9\\
0.85&88.38&78.9\\
0.9&86.68&79.5\\
0.95&84.10&80.7\\
\hline
\end{tabular}
\caption{Ablation study of cold start confidence threshold $CST$ on 8-shot GSM8K with LLaDA-1.5\citep{zhu2025llada}. }
\label{tab:aba-CST}
\end{table}


\section{Appendix: Comparing Against Naive Insertion Policy and Fixed Few-Shot Policy}\label{apd:naive-baseline}
We also evaluate DIP against a naive insertion policy that starts with minimal context and simply adds one example at each block boundary. As shown in Table \ref{tab:naive-ablation}, this naive policy yields suboptimal inference throughput. By indiscriminately increasing the prompt length at every step, the naive policy incurs unnecessary computational costs. DIP, by contrast, dynamically injects examples only when needed, demonstrating that strategically timed insertions are essential for maximizing throughput without sacrificing generation quality.

We also compared with a fixed few-shot policy on GSM8K in both the 5-shot and 8-shot settings. This apples-to-apples comparison compares fixed few-shot $ k$ vs. DIP with the same initial context ($n_0=k$), isolating exactly what dynamic insertion adds. As shown in Table \ref{tab:FS-ablation}, for every matched starting context, DIP either matches the fixed prompt or improves accuracy by using additional examples exactly when needed, at the expense of some inference throughput. However, DIP can balance accuracy and example usage (throughput) in DLM inference. This is exactly the flexibility we aim to demonstrate. Moreover, the best k is only knowable post hoc and does not transfer.

\begin{table}[t]
\centering
\begin{tabular}{|l|l|l|l|}
\hline
Shot & Method & Throughput & Acc \\
\hline
\multirow{3}{*}{5} 
 & Baseline   & 75.87\textcolor{blue}{(1.00$\times$)} & 77.7 \\
 & Naive ($n_0$=1)          & 81.32\textcolor{blue}{(1.07$\times$)} & 77.5 \\
 & DIP ($n_0$=1, RT=0)      & 87.08\textcolor{blue}{(1.15$\times$)} & 77.5 \\
\hline
\multirow{4}{*}{8}
 & Baseline   & 66.58 \textcolor{blue}{(1.00$\times$)} & 77.3 \\
 & Naive ($n_0$=1)          & 73.40 \textcolor{blue}{(1.10$\times$)} & 77.6 \\
 & DIP ($n_0$=1, RT=0)      & 87.80 \textcolor{blue}{(1.32$\times$)} & 76.2 \\
 & DIP ($n_0$=2, RT=0)      & 84.08 \textcolor{blue}{(1.26$\times$)} & 78.6 \\
\hline
\end{tabular}
\caption{Comparison of DIP against a naive monotonic-increasing policy on GSM8K with LLaDA-8B-Instruct.}
\label{tab:naive-ablation}
\end{table}

\begin{table*}[ht]
\centering
\begin{tabular}{|l|l|l|l|l|}
\hline
& Dataset & &   \multicolumn{2}{c|}{GSM8K} \\
\hline
Model & Method & Max \# of Example & Throughput & Acc \\ \hline
\multirow{8}{*}{LLaDA 8B-Instruct}&Baseline (fixed $k=1$)      & 5 & 95.67 & 73.0 \\
&DIP($n_0$=$1$)            & 5  & 87.08 & \textbf{77.5} \\ \cline{2-5}
&Baseline (fixed $k=2$)      & 5 & 90.05 & 77.1 \\
&DIP($n_0$=$2$)            & 5  & 83.85 & \textbf{77.1} \\ \cline{2-5}
&Baseline (fixed $k=1$)      & 8  & 95.76 & 74.4 \\
&DIP($n_0$=$1$)            & 8  & 90.24 & \textbf{76.4} \\ \cline{2-5}
&Baseline (fixed $k=2$)      & 8  & 89.35 & 74.5 \\
&DIP($n_0$=$2$)            & 8  & 84.08 & \textbf{78.6} \\
\hline
\multirow{8}{*}{LLaDA 1.5}&Baseline (fixed $k=1$)      & 5 & 100.40 & 76.6 \\
&DIP($n_0$=$1$)            & 5  & 86.55 & \textbf{78.9} \\ \cline{2-5}
&Baseline (fixed $k=2$)      & 5  & 94.18 & 78.6 \\
&DIP($n_0$=$2$)            & 5  & 85.63 & \textbf{79.5} \\ \cline{2-5}
&Baseline (fixed $k=1$)      & 8  & 98.01 & 77.1 \\
&DIP($n_0$=$1$)            & 8  & 87.73 & \textbf{79.2} \\ \cline{2-5}
&Baseline (fixed $k=2$)      & 8  & 93.12 & 79.8 \\
&DIP($n_0$=$2$)            & 8 & 82.92 & \textbf{80.3} \\
\hline

\end{tabular}
\caption{DIP's results comparing against the fixed-shot policy using LLaDA-1.5 and LLaDA-8B-Instruct.}
\label{tab:FS-ablation}
\end{table*}

\section{Appendix: Use of AI Assistants}
In this paper, we use AI to help with grammar and fluency. We also use AI for coding syntax assistance.


\end{document}